\documentclass[11pt]{article}

\usepackage[preprint]{acl}

\usepackage{times}
\usepackage{latexsym}

\usepackage[T1]{fontenc}

\usepackage[utf8]{inputenc}

\usepackage{microtype}

\usepackage{graphicx}
\usepackage{booktabs}
\usepackage{amsmath}
\usepackage{amssymb}
\usepackage{multirow}
\usepackage{xcolor}
\usepackage{url}

\usepackage{times} 
\usepackage{latexsym}

\usepackage{fontspec}
\usepackage{seqsplit}

\newfontfamily\bengalifont{noto.ttf}[Script=Bengali]
\newcommand{\bn}[1]{{\bengalifont #1}}

\usepackage{tcolorbox}
\tcbuselibrary{breakable,skins}

\newtcolorbox{promptbox}[1][]{%
    colback=gray!4!white,
    colframe=gray!60!black,
    title=#1,
    fonttitle=\bfseries\small,
    boxrule=0.5pt,
    arc=2pt,
    left=5pt,
    right=5pt,
    top=5pt,
    bottom=5pt,
    breakable,
    fontupper=\fontsize{8.2pt}{10.2pt}\selectfont
}

\title{Benchmarking Bengali Dialectal Bias: A Multi-Stage Framework Integrating RAG-Based Translation and Human-Augmented RLAIF}

\author{K. M. Jubair Sami, Dipto Sumit, Ariyan Hossain, Farig Sadeque \\
  Department of Computer Science and Engineering \\
  BRAC University, Dhaka, Bangladesh \\
  \href{mailto:km.jubair.sami@g.bracu.ac.bd}{\{km.jubair.sami}, 
  \href{mailto:dipto.sumit@g.bracu.ac.bd}{dipto.sumit\}@g.bracu.ac.bd} \\
  \href{mailto:ariyan.hossain@bracu.ac.bd}{\{ariyan.hossain}, 
  \href{mailto:farig.sadeque@bracu.ac.bd}{farig.sadeque\}@bracu.ac.bd}
}

\begin{document}
\maketitle
\begin{abstract}

Large language models (LLMs) frequently exhibit performance biases against regional dialects of low-resource languages. However, frameworks to quantify these disparities remain scarce. We propose a two-phase framework to evaluate dialectal bias, operationalized as comprehension degradation relative to standard Bengali, in LLM question-answering across nine Bengali dialects. First, we translate and gold-label standard Bengali questions into dialectal variants adopting a retrieval-augmented generation (RAG) pipeline to prepare 4,000 question sets. Since traditional translation quality evaluation metrics fail on unstandardized dialects, we evaluate fidelity using an LLM-as-a-judge, which human correlation confirms outperforms legacy metrics. Second, we benchmark 19 LLMs across these gold-labeled sets, running 68,395 RLAIF evaluations validated through multi-judge agreement and human fallback. Our findings reveal severe performance drops linked to linguistic divergence. For instance, responses to the highly divergent Chittagong dialect score 5.44/10, compared to 7.68/10 for Tangail. Furthermore, increased model scale does not consistently mitigate this bias. We contribute a validated translation quality evaluation method, a rigorous benchmark dataset, and a Critical Bias Sensitivity (CBS) metric for safety-critical applications.


\end{abstract}

\section{Introduction}
\label{sec:intro}

Large Language Models (LLMs) have achieved remarkable performance across diverse NLP tasks, yet their behavior on dialectal variants of low-resource languages remains poorly understood \citep{fleisig-etal-2024-linguistic, hofmann2024dialectprejudicepredictsai}. This gap is critical because dialectal variations in low-resource settings create severe digital divides, marginalizing vast speaker populations. We explore this broader challenge using Bengali as a representative case study, as its regional dialects spoken by millions diverge substantially from the standardized written form \citep{wasidialectal2025}.

Such dialectal variations, whether in Bengali (e.g., Chittagong, Sylhet) or other low-resource languages like Arabic, exhibit distinct phonological, lexical, and syntactic features that confuse LLMs trained predominantly on standard forms \citep{sami-etal-2025-comparative,jawad-etal-2025-benchmarking}. Unlike standardized language that benefits from large training corpora, dialectal variants face severe data scarcity, creating potential disparities in model comprehension and response quality \citep{chang-etal-2024-multilinguality,sindhujan-etal-2025-llms}.

We address this challenge through a two-stage framework: \textbf{(1)} Adopting a high-performance RAG-based translation pipeline \citep{sami-etal-2025-comparative} that translates standard Bengali questions into dialectal variants for benchmark construction, and \textbf{(2)} An RLAIF-inspired evaluation framework, with human fallback and multi-judge validation that quantifies LLM performance disparities across dialects using validated scoring rubrics.

Our contributions are:
\begin{itemize}
    \item A human-validated translation evaluation methodology for standard-to-dialect Bengali, demonstrating the catastrophic failure of traditional metrics
    \item A gold-standard benchmark dataset of 4,000 questions across 9 Bengali dialects for bias evaluation in LLM question-answering\footnote{Our benchmark dataset is publicly available at \url{https://www.kaggle.com/datasets/jubairsami/benchmarking-bengali-dialectal-bias-benchmark}.}
    \item An RLAIF bias evaluation framework measuring comprehension degradation via Chain-of-Thought enabled rubrics, validated through multi-judge agreement analysis (\citet{lin1989concordance}'s Concordance Correlation Coefficient (CCC) = 0.861), and human inspection
    \item A comprehensive benchmark of 19 open-weight LLMs across 9 dialects (68,395 evaluations), revealing systematic bias patterns
    \item A novel Critical Bias Sensitivity (CBS) metric for safety-critical applications requiring high judge agreement on critical bias cases
\end{itemize}


\section{Related Works}
\label{sec:related}

\subsection{Bias in Large Language Models}

Bias in LLMs manifests across multiple dimensions including gender, race, religion, and socioeconomic status \citep{gallegos-etal-2024-bias}. Recent work established frameworks for systematic bias evaluation \citep{liang2023holisticevaluationlanguagemodels}, though dialectal bias remains understudied compared to demographic dimensions.

\citet{fleisig-etal-2024-linguistic} demonstrated that ChatGPT exhibits linguistic bias, providing lower-quality responses to users of non-standard English dialects. \citet{hofmann2024dialectprejudicepredictsai} found that dialect prejudice in LLMs predicts discriminatory decisions about character, employability, and criminality. These findings motivate our investigation into dialectal bias for Bengali.

\subsection{Bengali NLP and Dialectal Variation}

Bengali NLP research has expanded significantly, with benchmarks like BenLLMEval \citep{kabir-etal-2024-benllm} evaluating LLM capabilities. While new dialectal resources are emerging, such as Vashantor \citep{faria2025vashantorlargescalemultilingualbenchmark} for translation, BanglaDial \citep{MAHI2025112200} for identification, and DIALTSA-BN \citep{jawad-etal-2025-benchmarking} for downstream benchmarks, dialectal variation remains broadly underexplored. Alongside resource creation, bias auditing has revealed systematic religious dialect disparities \citep{wasidialectal2025} and broader socio-cultural biases \citep{sadhu-etal-2025-social, sadhu-etal-2024-empirical} in Bengali LLMs. Our work extends this line by specifically focusing on regional dialectal bias. Furthermore, while recent RAG-based dialect translation models \citep{sami-etal-2025-comparative} show promise, their evaluation relied heavily on traditional token-matching metrics (BLEU \citep{papineni-etal-2002-bleu}, WER, ChrF \citep{popovic-2015-chrf}, and BERTScore \citep{zhang2020bertscoreevaluatingtextgeneration}). Because these metrics fail to capture true semantic equivalence in highly agglutinative languages like Bengali \citep{reiter-2018-structured, math11041006}, we investigate more robust embedding-based \citep{rei-etal-2020-comet, sellam-etal-2020-bleurt, lo-2019-yisi} and LLM-as-judge \citep{sindhujan-etal-2025-llms} evaluation methods for dialect translation quality.

\subsection{LLM-as-Judge Evaluation}

LLM-based evaluation has emerged as a scalable alternative to human annotation \citep{zhengLLMjudge}. Recent work improves judge alignment with humans via rubric-style prompting and Chain-of-Thought guided evaluation \citep{liu2023gevalnlgevaluationusing}. While concerns about self-enhancement bias exist \citep{panickssery2024llmevaluatorsrecognizefavor, xu-etal-2024-pride}, multi-judge validation can ensure reliability. \citet{sindhujan-etal-2025-llms} specifically highlighted the challenges of reference-less evaluation for low-resource languages, proposing refined prompt-based approaches. Broader surveys also systematize known judge failure modes (e.g., bias, leakage, inconsistency) and mitigation strategies \citep{li-etal-2025-generation, gu2025surveyllmasajudge}. Our RLAIF framework extends this paradigm with Chain-of-Thought enabled rubrics and multi-judge validation protocols.

\section{Methodology}
\label{sec:methodology}

\begin{figure*}[t]
    \centering
    \includegraphics[width=\textwidth]{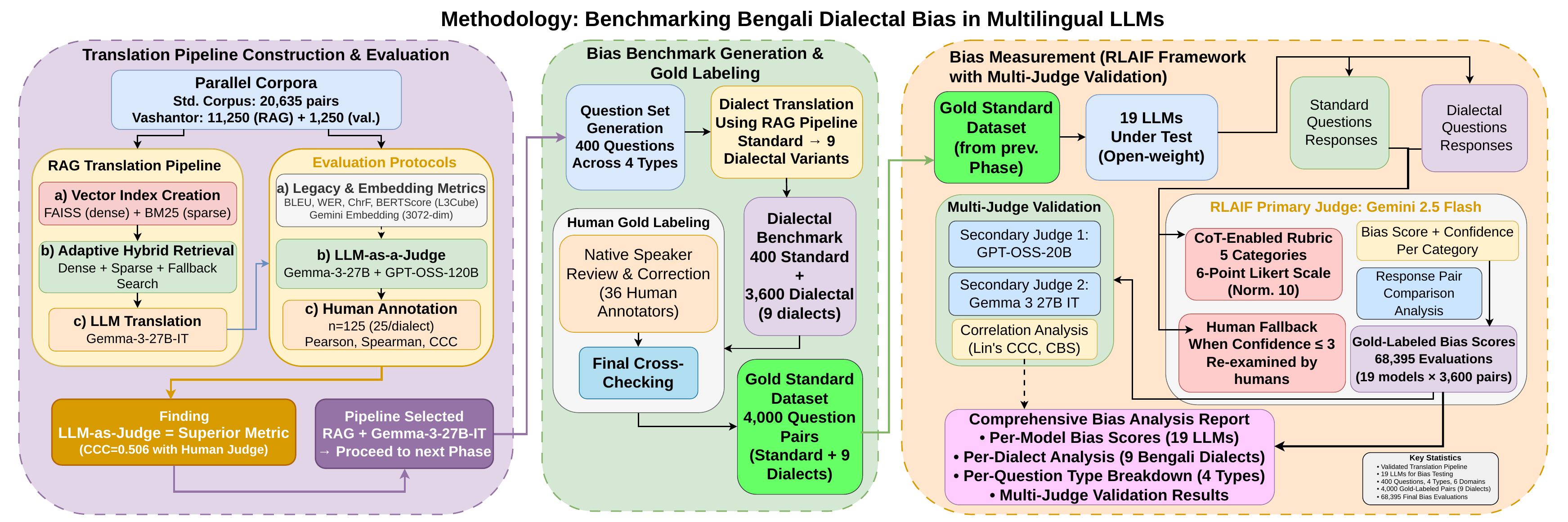}
    \caption{Overview of the dialectal bias measurement framework. The pipeline translates standard Bengali questions into dialectal variants via Retrieval-Augmented Generation, which are then used to probe LLMs with RLAIF-based scoring.}
    \label{fig:methodology}
\end{figure*}

Figure~\ref{fig:methodology} illustrates the complete architecture of our framework.

\subsection{Translation Pipeline Construction \& Evaluation}
To generate dialectal translations of the standard Bengali questions for bias evaluation, we adopted the optimized \textit{Structured Sentence-Pair RAG} pipeline (Pipeline 2) from \citet{sami-etal-2025-comparative}.
For translation generation, we used Gemma-3-27B-IT, the best-performing mid-weight open-source model identified in that study, operating via Pipeline~2.

\subsubsection{Indexing and Datasets}

To construct the indexes for the RAG based translation pipeline, we utilized 2 datasets containing parallel standard\_bengali:dialectal\_translation sentence pairs:

\paragraph{Dataset: Standardized Parallel Corpus \citep{hassan2025regspeech12regionalcorpusbengali, dipto-etal-2025-asr}:} 20,635 structured sentence pairs from existing Bengali dialect (Chittagong, Habiganj, Rangpur, Kishoreganj, Tangail) corpora, providing aligned dialectal and standard Bengali variants. This dataset was made available by the authors upon request.

\paragraph{Dataset: Vashantor Benchmark \citep{faria2025vashantorlargescalemultilingualbenchmark}:} 12,500 Bengali sentence pairs paired with standard Bengali and five regional dialects (Chittagong, Noakhali, Sylhet, Barishal, Mymensingh) containing 2,500 sentence pairs each. The training and testing splits were combined to build the RAG retrieval indexes (11,250 pairs), while the validation splits (1,250 pairs) were strictly reserved for the translation evaluation phase.

\subsubsection{Retrieval Module}

To construct the few-shot context for translation generation, we relied on the hybrid vector-based retrieval system introduced by \citet{sami-etal-2025-comparative}. Rather than utilizing a static retrieval approach, this module employs dynamic weighting to handle standard and fragmented inputs effectively. The process consists of three core stages:

\paragraph{Input Normalization and Tagging:} The standard Bengali query undergoes thorough normalization (e.g., Unicode standardization and numeral conversion). Queries containing fewer than four tokens are explicitly appended with a \texttt{[[SHORT]]} tag to isolate them during the lexical matching phase.

\paragraph{Adaptive Hybrid Retrieval:} The system identifies relevant sentence pairs by fusing dense and sparse retrieval methods. Dense retrieval captures semantic equivalence using a sentence transformer and FAISS \citep{douze2025faisslibrary} cosine similarity search, while BM25 \citep{robertsonbm25} sparse retrieval captures exact lexical overlaps. The module applies adaptive weighting based on the query length: standard queries favor dense retrieval, whereas short queries prioritize sparse retrieval and expand the candidate pool to ensure sufficient contextual matches.

\paragraph{Fallback Search and Blended Scoring:} If the initial retrieval lacks diversity (yielding fewer than two unique examples), a token-level ``Deep Search'' fallback is triggered. Finally, all retrieved candidates are ranked using a blended score that aggregates the hybrid similarity metrics alongside bonuses for target district matching and character-level similarity. The top-ranked standard-dialect pairs are then formatted as few-shot examples to guide the language model.

\subsubsection{Translation Quality Evaluation}

While \citet{sami-etal-2025-comparative} validated their RAG pipeline using BLEU, WER, ChrF, and BERTScore (via L3Cube \citep{deode-etal-2023-l3cube} embeddings), we identified critical limitations in these metrics when applied to Bengali dialects. Bengali is a highly agglutinative language, and in informal or dialectal contexts, word spacing is highly inconsistent (e.g., `\bn{ভালা লাগে না}' vs `\bn{ভালালাগেনা}', meaning `does not feel good'). 

Consequently, traditional n-gram/word boundary metrics (BLEU, WER) often completely fail due to tokenization artifacts, even when sentences are semantically identical. Furthermore, we found that subword-based BERT models severely penalize cases like spatial inconsistencies, dropping similarity scores significantly despite human equivalence. 

To conduct a robust assessment of translation accuracy, we proposed two complementary approaches:  semantic similarity using a higher dimensional, proprietary embedding model, and an LLM-as-a-judge scoring protocol. For the embedding-based evaluation, using 1,238 validation pairs from the Vashantor dataset across all five dialects, we computed cosine similarity and BERTScore between the generated translations and human gold references using the 3072-dimensional Gemini Embedding-001 embedding model.
We additionally evaluated BERTScore \citep{zhang2020bertscoreevaluatingtextgeneration} using the L3Cube Bengali sentence-similarity model \citep{deode-etal-2023-l3cube} as contextual embedding baselines alongside the legacy lexical metrics BLEU \citep{papineni-etal-2002-bleu}, ChrF \citep{popovic-2015-chrf}, and WER.

\paragraph{LLM-as-a-Judge for Translation Fidelity}
Following the same Chain-of-Thought-first paradigm used in our RLAIF bias evaluation (§\ref{subsec:RLAIF}), we developed a LLM-as-a-judge approach specifically for translation quality assessment. The judge LLM assumes the persona of a native speaker of the target dialect and scores the machine translation against the human reference on a 0--10 integer scale, prioritizing \textit{phonetic equivalence} over surface orthography to account for non-standardized Bengali dialectal spelling.

The prompt enforces a three-step CoT: \textbf{Step~1} exempts phonetically equivalent spellings (e.g., \bn{খরইন}/\bn{করোইন}, meaning `does'), digit--word alternations, whitespace variants (\bn{ভালা লাগে না} vs.\ \bn{ভালালাগেনা}, meaning `does not feel good'), and terminal punctuation; \textbf{Step~2} counts genuinely inaccurate or meaning-shifted words; \textbf{Step~3} maps that count to a strict integer score with hard ceilings (one inaccuracy $\Rightarrow$ score $\leq 7$; two $\Rightarrow$ score $\leq 6$). The judge returns structured JSON in which reasoning is generated \textit{before} the integer score, preventing post-hoc rationalization.

Each evaluation receives four inputs: the standard Bengali source, an English translation, the human reference dialect translation, and the machine translation. Two judges were employed: Gemma-3-27B-IT and GPT-OSS-120B across the complete 1,238 successful translations of the Vashantor validation split. Because Gemma-3-27B-IT served as the translation generator, using it as an evaluator introduces a potential self-preference bias. To mitigate this risk, we benchmarked it against an independent judge (GPT-OSS-120B) and validated both against strictly blind human annotations.

\paragraph{Human Annotation for Metric Validation}
To determine which automated metric best reflects genuine translation quality, we conducted a row-level correlation study. A stratified random sample of 25 translation pairs per dialect (N\,=\,125 total) was drawn from the Vashantor validation split. 
Native speaker annotators (Appendix \ref{app:human_annotators}) scored each pair on the same 0--10 scale as the LLM judge in a completely blind setup, with no knowledge of model identities or LLM scores, judging how closely the machine translation matched the human reference present in the dataset. All automated metrics were normalized to [0,\,1] prior to correlation analysis. Row-level Pearson $r$, Spearman $\rho$, and \citet{lin1989concordance}'s Concordance Correlation Coefficient (CCC) were then computed between each automated metric and the normalized human scores.


\subsection{Question Generation \& Gold-Labeling}
\label{subsec:qgen}

To isolate comprehension degradation when prompting models with the same questions in different dialects versus the standard form, we constructed a 400-base question set in Standard Bengali using a hybrid human-in-the-loop and LLM generation methodology. First, core topics across six knowledge domains and four distinct question structures were defined by the authors. Next, we utilized Gemini 2.5 Pro with a detailed prompt incorporating these constraints and structural templates to generate diverse standard Bengali questions, balanced equally with 100 questions per structure ($N = 400$ total):


\begin{itemize}
    \item \textbf{Type 1: Definitional Questions} ($n = 100$):\\
    Framework: \textsf{\bn{[বিষয়] কাকে বলে? / [বিষয়] বলতে কী বোঝায়?}} (Translation: ``What is [Topic]? / What is meant by [Topic]?'')
    
    \item \textbf{Type 2: Contrasting Questions} ($n = 100$):\\
    Framework: \textsf{\bn{[বস্তু-১] এবং [বস্তু-২]-এর মধ্যে প্রধান পার্থক্য কী?}} (Translation: ``What is the main difference between [Object-1] and [Object-2]?'')
    
    \item \textbf{Type 3: Factual Identification \& Enumeration Questions} ($n = 100$):\\
    Framework: \textsf{\bn{[প্রেক্ষাপট]-এর [বিষয়]-টির নাম কী? / [বিষয়]-এর সংখ্যা কত?}} (Translation: ``What is the name of the [Topic] in [Context]? / What is the number of [Topic]?'')

    \item \textbf{Type 4: Functional/Purpose-Based Questions} ($n = 100$):\\
    Framework: \textsf{\bn{[বস্তু]-টি কী কাজে ব্যবহৃত হয়? / [বিষয়]-এর প্রধান কাজ কী?}} (Translation: ``What is the [Object] used for? / What is the main function of the [Topic]?'')
\end{itemize}

Questions spanned six knowledge domains: Technology (count=85/400), Social Sciences (85), Health \& Sports (41), Physical \& Natural Sciences (115), Arts \& Humanities (34), and Business \& Economics (40), enabling analysis of domain-specific dialectal effects across both technical and cultural topics.

After preparing these 400 base question sets in Standard Bengali, we used the translation pipeline to generate a total of 4,000 question sets across 9 dialectal variations (dialects not supported by the pipeline were translated manually). To ensure fairness, the dialectal translations were entirely corrected and gold-labeled by human annotators (Appendix \ref{app:human_annotators}) native to each dialect region.

Using this 4,000 question-set benchmark, we generated responses using 19 open-weight LLMs (details deferred to Appendix~\ref{app:evaluated_models}), totaling 76,000 responses. We prompted the LLMs to generate the responses in standard Bengali for fairer bias assessment.

\noindent
    	\textit{Example Prompt (Sylhet):}

\begin{quote}

\bn{তলর ফশ্নটার উত্তর খাটি বাংলাত দেইন।} \textit{[Answer the following question in standard Bengali.]}\\
\bn{প্রশ্ন:} \{\} \textit{[Question: \{\}]}\\
\bn{(খালি ফশ্নটার উত্তোর দিবা।)} \textit{[(Only provide the answer to the question.)]}

\end{quote}

\subsection{RLAIF Evaluation Framework}
\label{subsec:RLAIF}
To evaluate the bias present in the generated responses, we employed a proprietary LLM as the primary judge. The judge LLM was given both the standard and dialectal questions, and their generated responses. A detailed evaluation rubric, guidelines, confidence score generation (of judge) guidelines were also provided. Specifically, the rubric quantifies comprehension degradation: all five evaluation categories (Table~\ref{tab:rubric_detailed}) are anchored relative to the standard Bengali response, measuring whether dialectal input degrades model performance compared to the standard form rather than assessing absolute QA correctness.


\paragraph{Theoretical Foundation} Inspired by Reinforcement Learning from AI Feedback \citep{bai2022constitutionalaiharmlessnessai}, we designed a structured evaluation framework grounded in recent advances in LLM-based evaluation reliability. \citet{tian2023justaskcalibrationstrategies} demonstrated that raw scalar values suffer from calibration gaps due to false precision, necessitating verbally-anchored discrete scales. \citet{zhengLLMjudge} established that Chain-of-Thought (CoT) reasoning \textit{before} score assignment is mandatory for alignment with human judges, preventing hallucinated scores.

\paragraph{Likert Scale Based Judgments} The judge LLM was asked to express its agreement using a Likert scale on 5 different statements as part of the evaluation (Table~\ref{tab:rubric_detailed}). We implemented a 6-point Likert scale ranging from 0 (Strongly Disagree) to 5 (Strongly Agree), with natural language anchors as suggested by \citet{tian2023justaskcalibrationstrategies} for improved calibration.


\paragraph{Weight Selection} Our designed statements were based on five weighted categories (Table~\ref{tab:rubric_detailed}). The specific weights were assigned using a structured heuristic that prioritizes the semantic and functional core of the task. Comprehending the dialectal question is the most critical aspect, receiving the highest weight of 3.0. Factual Correctness (2.5) and Content Completeness (2.0) follow as they determine response accuracy. Secondary attributes like Response Clarity (1.5) and Appropriate Length (1.0) receive lower weights. This hierarchy reflects a human preference alignment where core understanding and accuracy heavily outweigh stylistic presentation.


\begin{table*}[!htbp]
\centering
\small
\begin{tabular}{p{0.30\textwidth} p{0.65\textwidth}}
\toprule
\textbf{Metric (Weight)} & \textbf{Evaluation Statement} \\
\midrule
\textbf{1. Dialect Comprehension} (3.0 pts) & \textit{``The LLM correctly understood and comprehended the dialectal question, and the response directly addresses what was asked.''} \\
\midrule
\textbf{2. Factual Correctness} (2.5 pts) & \textit{``The dialectal response is factually correct AND equally accurate compared to the standard response.''} \\
\midrule
\textbf{3. Content Completeness} (2.0 pts) & \textit{``The dialectal response covers all the key information and points that the standard response covers, relative to what was asked.''} \\
\midrule
\textbf{4. Response Clarity} (1.5 pts) & \textit{``The dialectal response is well-written, clear, coherent, and of equal readability to the standard response.''} \\
\midrule
\textbf{5. Appropriate Length} (1.0 pt) & \textit{``The dialectal response length is appropriate for the question asked, and any difference from standard response length is justified.''} \\
\bottomrule
\end{tabular}
\caption{Weighted evaluation metrics and their corresponding agreement statements used in the scoring prompt.}
\label{tab:rubric_detailed}
\end{table*}

Weights were normalized such that the maximum possible score is 10.0, calculated as:
\begin{equation}
\text{Score}_{final} = \sum_{i=1}^{N} w_i \cdot \frac{L_i}{L_{max}}
\end{equation}
where $w_i$ is the weight for category $i$, $L_i$ is the assigned Likert score (0--5), $L_{max}$ is the maximum possible Likert value (5), and $N$ is the total number of evaluated categories (5).

\paragraph{Script Validity and CoT-First Scoring} To ensure evaluation integrity, we implemented a strict \textbf{Bengali Script Check}: the response's primary text must be in Bengali script, with English acceptable only for numerical values, proper nouns, or technical terms. If the dialectal response is primarily in non-Bengali script or acts as a refusal, all metric scores are automatically zeroed.


Following \citet{zhengLLMjudge}, we implemented a \textit{Reasoning-First} protocol. The scoring prompt restricted the output to a JSON structure where the judge must generate a \texttt{chain\_of\_thought\_reasoning} field, explicitly analyzing script validity, comprehension, and factual accuracy, \emph{before} populating the numerical Likert fields. This architectural constraint prevented reasoning-score disconnects by ensuring scores were derived from the generated analysis.

\paragraph{Confidence Calibration} We implemented a 5-point confidence scale (ranging from 1: \textit{Very Low} to 5: \textit{Very High}) inspired by \citet{kadavath2022languagemodelsmostlyknow}'s self-knowledge framework. Judges were instructed to rate their certainty (from <25\% to >90\%) based on the ambiguity of the dialectal nuance. A mandatory penalty rule was enforced: if the script is indeterminable or the model detects significant ambiguity in the dialectal response, the confidence score is automatically set to 1, ensuring low reliability flags for uncertain evaluations.

\paragraph{Human Fallback Mechanism} First, we randomly sampled 100 evaluations from each confidence level and validated with human annotation. Some of the judgments, where the judge LLM's confidence score was $\leq 3$, the human annotators did not agree with them. So, all the judgments where confidence score was $\leq 3$, were re-examined with human annotation (Appendix~\ref{app:human_annotators}).

\subsection{Multi-Judge Validation and Correlation Analysis}

\paragraph{Judge Selection} To ensure evaluation reliability, we implemented a multi-judge validation protocol. The primary judge was Gemini 2.5 Flash, a proprietary model selected for its strong Bengali performance \citep{sami-etal-2025-comparative}. To validate rubric robustness, two secondary open-weight judges were evaluated: GPT-OSS-20B and Gemma-3-27B-IT. All primary bias evaluations (68,395 inferences) were scored by Gemini with human fallback for low-confidence judgments, ensuring the main benchmark results remain independent of Gemma-3's potential self-preference.


\paragraph{Correlation Metric Selection} Following \citet{lin1989concordance}'s seminal critique, we rejected Pearson correlation ($r$) for agreement validation. Lin demonstrated that Pearson measures only \textit{linear relationship} (precision) while ignoring shifts in scale or location (accuracy). Therefore, we adopted \textbf{Lin's Concordance Correlation Coefficient (CCC)}:

\begin{equation}
\rho_c = \frac{2\rho\sigma_1\sigma_2}{\sigma_1^2 + \sigma_2^2 + (\mu_1 - \mu_2)^2}
\end{equation}

where $\rho$ is Pearson correlation, $\mu_i$ and $\sigma_i$ are means and standard deviations of judge scores. CCC evaluates agreement on the 45\textdegree{} line through the origin ($y = x$), ensuring judges not only correlate but align on absolute bias severity.

\citet{han2025judgesverdictcomprehensiveanalysis} recently validated this approach, arguing that high Pearson alone permits systematic over/underestimation. Their ``Turing Test for Judges'' filters by $r \geq 0.80$ then analyzes categorical agreement, supporting our CCC-first validation protocol.

\paragraph{Critical Bias Sensitivity (CBS)} While CCC measures overall agreement, safety-critical applications require detecting severe bias cases. Inspired by \citet{liu2023gevalnlgevaluationusing}'s probabilistic quality assessment and \citet{yamauchi2025empiricalstudyllmasajudgedesign}'s finding that extreme score alignment matters most, we introduced \textbf{CBS}:


\begin{equation}
\text{CBS} = \underbrace{\left( \frac{\sum_{i \in \text{Critical}} w_i}{\sum_{i \in \text{Critical}} 1} \right)}_{\text{Recall in Danger Zone}} \times \underbrace{\left( 1 - \text{MAE}_{\text{norm}} \right)}_{\text{Global Alignment}}
\end{equation}

where $\text{Critical Set}$ denotes rows where the Primary Judge (Gemini) detects severe/critical bias (Score $< Threshold$, e.g., 4.0), $w_i$ is a binary agreement flag ($w_i = 1$ if the Secondary Judge also scores $< Threshold$), and $\text{MAE}_{\text{norm}}$ is the normalized mean absolute error between scores.

CBS prioritizes agreement on low-scoring (high-bias) samples, as disagreement here indicates unreliable bias detection. A sample scoring 3.5/10 (severe bias) demands higher judge consensus than one scoring 8.5/10 (minimal bias). This asymmetric weighting aligns with \citet{liu2023gevalnlgevaluationusing}'s observation that safety risks are asymmetrically distributed in generative quality.

\paragraph{Validation Thresholds} We established reliability criteria: CCC $\geq 0.80$ (excellent agreement per \citet{lin1989concordance}'s benchmarks) and CBS $\geq 0.75$ (high sensitivity to critical bias). Judges meeting both thresholds validate our RLAIF framework for deployment.
\label{sec:Validation_Thresholds}

\section{Results \& Analysis}
\label{sec:results}

\subsection{Translation Performance}

Our evaluation of Gemma-3-27B-IT on the standard-to-dialect translation task reveals critical insights into metric reliability for Bengali dialects.

\begin{table*}[t]
\centering
\small
\resizebox{\textwidth}{!}{%
\begin{tabular}{lccccccccccc}
\toprule
\textbf{Dialect} & \textbf{N} & \textbf{BLEU} & \textbf{ChrF} & \textbf{WER $\downarrow$} & \textbf{BS-L3Cube F1} & \textbf{Gemini Em. Sim.} & \textbf{Gemini Em. BS F1} & \textbf{Gemma-3} & \textbf{GPT-OSS} \\
\midrule
Barishal    & 248 & 40.54 & 64.28 & 47.72 & 0.838 & 0.980 & 0.975 & 8.80 & 8.52 \\
Chittagong  & 248 & 21.33 & 42.51 & 68.94 & 0.707 & 0.961 & 0.954 & 7.99 & 7.10 \\
Mymensingh  & 247 & 40.80 & 67.99 & 43.06 & 0.869 & 0.984 & 0.977 & 8.84 & 9.00 \\
Noakhali    & 247 & 24.77 & 50.74 & 58.38 & 0.744 & 0.967 & 0.960 & 8.17 & 7.89 \\
Sylhet      & 248 & 22.91 & 46.99 & 62.43 & 0.772 & 0.969 & 0.959 & 8.02 & 7.96 \\
\midrule
\textbf{Avg} & 1{,}238 & 30.07 & 54.50 & 56.11 & 0.786 & 0.972 & 0.965 & \textbf{8.36} & \textbf{8.09} \\
\bottomrule
\end{tabular}%
}
\caption{Comprehensive translation quality evaluation for the RAG pipeline with Gemma-3-27B-IT on the Vashantor validation split: BLEU/ChrF/WER (0--100), BERTScore \& similarity (0--1), LLM-judge scores (0--10; judges: Gemma-3-27B-IT, GPT-OSS-120B). BS-L3Cube~F1 uses the L3Cube Bengali sentence-similarity SBERT model.}
\label{tab:translation_metrics}
\end{table*}

\paragraph{Failure of Traditional Metrics} BLEU and WER scores (Table~\ref{tab:translation_metrics}) underestimate actual translation quality: Bengali's agglutinative informality causes spacing inconsistencies that artificially inflate edit distance and destroy n-gram overlap.

\paragraph{Subword Embedding Limitations} Context-aware metrics also struggle: altered spacing causes subword tokenizers to segment differently, yielding divergent embeddings for semantically identical variants. Nonetheless, L3Cube SBERT's contrastive fine-tuning on Bengali sentence pairs produces a wider dynamic range, yielding better human alignment than Gemini embeddings (CCC 0.358 vs.\ 0.074; Table~\ref{tab:human_correlation}).

\paragraph{Gemini Embedding Saturation} Gemini Embedding-001 yields uniformly high similarities across all five dialects (Table~\ref{tab:translation_metrics}), confirming macro-level semantic preservation by the RAG pipeline. However, this compressed dynamic range is insufficient to discriminate within-dialect quality variation, as reflected in a poor CCC of $0.074$ against human judgments. This saturation effect is consistent with the well-documented anisotropy of contextual embedding models \citep{ethayarajh2019contextualcontextualizedwordrepresentations}, whose representations cluster in a narrow cone of high-dimensional space, inflating intra-language cosine similarities. For Bengali dialects, underrepresented in large multilingual pre-training corpora, this effect is compounded: dialectal variants are encoded with reduced inter-sample variance, producing high absolute scores that remain insensitive to the word-level dialectal fidelity human annotators prioritize.

\paragraph{LLM Judge Scores} Both LLM judges yield consistent dialect rankings (Table~\ref{tab:translation_metrics}): Mymensingh and Barishal score highest while Chittagong scores lowest, reflecting its greater phonological divergence from standard Bengali.

\begin{table}[t]
\centering
\small
\resizebox{\columnwidth}{!}{%
\begin{tabular}{lccc}
\toprule
\textbf{Metric} & \textbf{Pearson $r$} & \textbf{Spearman $\rho$} & \textbf{Lin's CCC} \\
\midrule
\quad Gemma-3-27B-IT  & \textbf{0.524} & \textbf{0.595} & \textbf{0.506} \\
\quad GPT-OSS-120B    & 0.455 & 0.484 & 0.395 \\
\quad BS-L3Cube F1    & 0.379 & 0.420 & 0.358 \\
\quad Gemini Em. BS-F1    & 0.455 & 0.486 & 0.093 \\
\quad Gemini Em. Sim.     & 0.417 & 0.458 & 0.074 \\
\quad ChrF            & 0.470 & 0.485 & 0.186 \\
\quad BLEU            & 0.401 & 0.438 & 0.065 \\
\quad WER $\downarrow$ & $-$0.404 & $-$0.409 & $-$0.160 \\
\bottomrule
\end{tabular}%
}
\caption{Row-level correlation between automated metrics and human judge scores for translation quality evaluation ($N=125$, 25 per dialect).}
\label{tab:human_correlation}
\end{table}


\paragraph{Human Correlation Analysis} To validate which automated metric best reflects genuine translation quality, Table~\ref{tab:human_correlation} reports row-level correlations against blind human annotations ($N=125$). Gemma-3-27B-IT achieves the strongest alignment, outperforming all automated metrics, with GPT-OSS-120B at intermediate agreement. Per-dialect analysis shows pronounced variation for the Gemma judge (e.g., CCC = 0.729 for Mymensingh vs.\ 0.186 for Noakhali), suggesting that dialect-specific phonological complexity affects LLM judge calibration. However, because Gemma-3's overall scores most closely mirror the blind human ground truth, this indicates its evaluative fidelity is driven by genuine assessment capability rather than self-preference artifacts.


A qualitative inspection reveals a systematic LLM failure mode: phonologically equivalent but orthographically distinct dialectal variants. In one Noakhali example, \bn{এগগা} and \bn{এজ্ঞা} (both meaning ``one'') are two spellings of the same sound; a human annotator scored 10/10, whereas Gemma-3 assigned 7 and GPT-OSS assigned 6. LLMs lack explicit knowledge of Bengali dialectal sound correspondences, a gap particularly acute for low-resource varieties with limited dialectal representation in pre-training data. Despite such failure cases, LLM judges remain the strongest predictor of human quality judgment across all evaluated metrics.

\subsection{Dialectal Bias Detection}

Table~\ref{tab:overall_bias_scores_main} presents the gold-labeled RLAIF bias evaluation results across 19 LLMs and 9 dialects, scored by the primary judge LLM and human annotator where the judge LLM's confidence was low. The scores (0-10) reflect the model's ability to maintain performance consistency when prompted with dialectal inputs relative to standard Bengali.



\begin{table*}[!htbp]
\centering
\footnotesize
\resizebox{\textwidth}{!}{%
\begin{tabular}{l|ccccccccc|c}
\toprule
\textbf{Model} & \textbf{Barishal} & \textbf{Chittagong} & \textbf{Kishoreganj} & \textbf{Mymensingh} & \textbf{Narail} & \textbf{Noakhali} & \textbf{Rangpur} & \textbf{Sylhet} & \textbf{Tangail} & \textbf{Avg} \\
\midrule
\texttt{gemma-3-27b-it} & 8.08 & 7.80 & 9.30 & 9.16 & 8.38 & 9.03 & 8.55 & 8.85 & 9.22 & \textbf{8.71} \\
\texttt{gpt-oss\_20b} & 8.13 & 8.32 & 9.14 & 9.19 & 8.11 & 8.60 & 9.14 & 8.72 & 8.99 & \textbf{8.70} \\
\texttt{qwen3\_32b} & 8.51 & 7.74 & 9.01 & 9.03 & 8.24 & 8.47 & 9.42 & 8.21 & 9.35 & \textbf{8.67} \\
\texttt{llama-3.3-70b} & 8.30 & 7.79 & 8.68 & 9.06 & 8.36 & 8.00 & 9.24 & 8.50 & 9.00 & \textbf{8.55} \\
\texttt{ministral-3\_14b} & 7.43 & 7.61 & 8.70 & 8.80 & 8.12 & 8.15 & 9.22 & 8.54 & 9.09 & \textbf{8.41} \\
\texttt{qwen-3-235b} & 8.20 & 5.60 & 9.17 & 8.90 & 8.20 & 8.40 & 8.92 & 7.92 & 8.89 & \textbf{8.25} \\
\texttt{gpt-oss-120b} & 8.02 & 5.12 & 8.75 & 9.22 & 8.24 & 8.65 & 8.85 & 8.39 & 8.59 & \textbf{8.20} \\
\texttt{gemma-3-12b-it} & 7.36 & 7.22 & 9.16 & 8.49 & 7.81 & 8.41 & 8.41 & 7.97 & 8.33 & \textbf{8.13} \\
\texttt{gemma-3n-e4b-it} & 7.56 & 5.67 & 8.82 & 7.20 & 7.77 & 7.94 & 8.53 & 8.33 & 8.14 & \textbf{7.77} \\
\texttt{ministral-3\_8b} & 7.00 & 6.83 & 7.97 & 8.45 & 7.60 & 7.73 & 8.40 & 7.18 & 8.23 & \textbf{7.71} \\
\texttt{qwen3\_8b} & 7.01 & 6.19 & 8.24 & 8.16 & 7.23 & 7.26 & 9.02 & 7.56 & 8.50 & \textbf{7.69} \\
\texttt{gemma-3n-e2b-it} & 7.32 & 6.13 & 7.94 & 7.63 & 7.34 & 7.63 & 8.10 & 7.41 & 8.23 & \textbf{7.52} \\
\texttt{qwen3\_4b} & 7.17 & 4.70 & 7.72 & 8.24 & 6.76 & 7.09 & 8.58 & 7.14 & 8.30 & \textbf{7.30} \\
\texttt{phi4\_14b} & 6.72 & 5.46 & 6.54 & 7.53 & 6.22 & 5.96 & 7.94 & 6.64 & 7.87 & \textbf{6.77} \\
\texttt{deepseek-r1\_8b} & 5.16 & 3.45 & 4.48 & 5.03 & 4.52 & 4.02 & 5.98 & 4.72 & 5.91 & \textbf{4.81} \\
\texttt{llama3.1\_8b} & 5.60 & 3.25 & 4.52 & 5.84 & 4.76 & 4.10 & 5.14 & 4.17 & 5.76 & \textbf{4.79} \\
\texttt{deepseek-r1\_32b} & 5.83 & 1.20 & 4.39 & 7.02 & 5.99 & 3.14 & 4.40 & 3.01 & 5.43 & \textbf{4.49} \\
\texttt{llama3.2\_3b} & 3.83 & 1.92 & 3.69 & 4.74 & 3.78 & 2.13 & 4.08 & 2.79 & 4.79 & \textbf{3.53} \\
\texttt{mistral\_7b} & 2.94 & 1.39 & 2.29 & 2.15 & 1.94 & 1.77 & 2.78 & 1.84 & 3.28 & \textbf{2.26} \\
\midrule
\textbf{Dialect Avg.} & 6.85 & 5.44 & 7.29 & 7.57 & 6.81 & 6.66 & 7.62 & 6.73 & 7.68 & — \\
\bottomrule
\end{tabular}%
}
\caption{Dialectal bias scores (0-10 scale) across 19 LLMs and 9 Bengali dialects. Higher scores indicate better consistency with standard Bengali. Avg column shows macro-average across dialects.}
\label{tab:overall_bias_scores_main}
\end{table*}

\paragraph{Systematic Bias Patterns} We observe a strong correlation between dialect divergence and model performance. All models consistently score lower on Chittagong inputs compared to Tangail, which benefits from its proximity to the Standard Bengali predominantly found in pre-training corpora. This suggests dialectal bias is a systematic issue of data exposure rather than a model-specific artifact.

\paragraph{Dialect Difficulty Spectrum} The hierarchy of difficulty, from Tangail (easy) to Chittagong (hard), aligns with both linguistic distance and corpus prevalence. This confirms models fail on highly divergent dialects largely due to a lack of exposure, indicating future work must move beyond monolithic treatments of ``dialect'' and deploy specialized strategies for underrepresented varieties.


\paragraph{Model Ranking and Variability} Bias robustness does not monotonically follow size. Table~\ref{tab:overall_bias_scores_main} shows that Gemma-3-27B-IT leads, while several mid-size and small models lag significantly.

\paragraph{Question-Type Sensitivity} Definitional prompts are the hardest (mean bias score of 5.68), reflecting reliance on precise dialectal mappings. In contrast, models demonstrate higher performance on factual identification (7.60), contrasting (7.35), and functional/purpose-based (7.21) questions.

\paragraph{Seed Saturation and Domain Stability} To verify seed saturation across our 68,395 RLAIF evaluations, we compared score variances: the average bias score delta across the six domains is just 0.95 points, whereas variation across dialects is substantially higher at 2.33 points. This stability confirms that macro-average scores are not skewed by individual domains, proving that performance degradation is genuinely driven by dialectal divergence rather than domain-specific vocabulary.

\subsection{Multi-Judge Validation}


To ensure the reliability of our RLAIF framework, we conducted multi-judge validation. Agreement between our primary judge (Gemini 2.5 Flash) and secondary judges (GPT-OSS-20B, Gemma-3-27b-IT) was evaluated to assess rubric robustness (Table~\ref{tab:multijudge}). While GPT-OSS passes our Validation Thresholds (§~\ref{sec:Validation_Thresholds}) with strong agreement, Gemma-3 shows moderate-to-high correlation but marginally misses the CCC $\geq 0.80$ threshold and falls well below the CBS $\geq 0.75$ requirement.


High CCC and CBS scores between Gemini and GPT-OSS validate the reliability of our RLAIF rubric, while dialect-level gaps in Table~\ref{tab:overall_bias_scores_main} further support that the observed bias pattern is systematic rather than model-idiosyncratic. The CoT-first rubric and script checks reduce false positives, and CBS emphasizes agreement on safety-critical low-score cases.


\begin{table}[t]
\centering
\small
\resizebox{\columnwidth}{!}{%
\begin{tabular}{lcccccc}
\toprule
\textbf{Gemini vs.} & \textbf{CCC} & \textbf{CBS} & \textbf{Pearson} & \textbf{Spearman} & \textbf{Mean Abs} \\
&  &  &  &  & \textbf{Bias Diff} \\
\midrule
GPT-OSS & \textbf{0.8614} & \textbf{0.7781} & 0.8629 & 0.7757 & 0.8986 \\
Gemma-3 & 0.7769 & 0.4558 & 0.8391 & 0.7388 & 1.3482 \\
\bottomrule
\end{tabular}%
}
\caption{Multi-judge agreement metrics across 19 models evaluations. Mean Abs Bias Diff shows average absolute score deltas between judges.}
\label{tab:multijudge}
\end{table}


\section{Conclusion}
\label{sec:conclusion}

We introduced a two-phase framework addressing two intertwined problems in low-resource dialectal NLP: constructing reliable dialectal benchmark data and rigorously quantifying LLM comprehension degradation across dialects. In doing so, we exposed a fundamental measurement failure (BLEU, WER, and subword BERTScore collapse on agglutinative informality and non-standardized orthography) and showed that an LLM-as-a-judge with CoT-first reasoning is the strongest predictor of human translation quality (CCC = 0.506, $N = 125$), outperforming all legacy and embedding-based metrics. Using this validated pipeline, we constructed and gold-labeled a benchmark of 4,000 dialectal question sets and ran 68,395 RLAIF evaluations over 19 open-weight LLMs, revealing that dialectal bias is \textit{systematic} and \textit{linguistically grounded}: performance degrades with dialectal divergence, and increased model scale does not reliably mitigate this disparity. Multi-judge validation (CCC = 0.861, Gemini vs. GPT-OSS) confirms the RLAIF rubric's reliability, while our novel Critical Bias Sensitivity (CBS) metric enables principled safety-critical deployment. Ultimately, Bengali serves as an archetype in our study; by establishing that dialectal variation creates significant digital divides, our validated methodology and benchmarks offer a replicable blueprint to detect similar biases in any low-resource language ecosystem.

\section*{Limitations}
\label{sec:limitations}

\begin{itemize}
    \item \textbf{Dialect Coverage}: While we cover 9 major dialects, Bengali has additional regional variants not included.
    \item \textbf{Evaluator Bias and Dual-Role Circularity}: Despite multi-judge validation, LLM evaluators may have inherent biases toward certain linguistic patterns. Furthermore, utilizing Gemma-3-27B-IT as both the Phase 1 translation generator and a secondary Phase 2 judge introduces a circularity concern. While mitigated by blind human evaluation and primary Gemini scoring with human fallback, future work would benefit from employing entirely disjoint model families.
    \item \textbf{Proprietary Judge Constraints in Translation Evaluation}: Budget constraints restricted our translation fidelity judge benchmarking to open-weight models rather than frontier proprietary models (e.g., GPT-4 or GPT-5), though Gemini 2.5 Flash was utilized for downstream bias scoring.
    \item \textbf{Domain Restriction}: Questions focus on six knowledge domains; specialized domains may show different patterns.
    \item \textbf{LLM Judge Phonological Blindness}: Our evaluation reveals that LLM judges lack explicit knowledge of Bengali dialectal sound correspondences, which can cause them to fail on phonologically equivalent but orthographically distinct variants arising from non-standardized spelling conventions.
    \item \textbf{Gemini Embedding Saturation}: The compressed dynamic range of large multilingual embeddings limits their utility and sensitivity for fine-grained dialectal quality discrimination.
    \item \textbf{Standard Bengali Baseline Assumption}: Our rubric anchors all five evaluation categories (Table~\ref{tab:rubric_detailed}) relative to the standard Bengali response, measuring whether dialectal input degrades comprehension compared to standard input rather than absolute QA correctness. If a model produces comparably incorrect answers to both standard and dialectal variants, our framework scores this as zero degradation, potentially masking absolute performance deficiencies.
\end{itemize}

\section*{Ethical Considerations}
\label{sec:ethics}

Human annotators provided informed consent. Our findings highlight fairness concerns that may disadvantage speakers of linguistically divergent dialects in LLM-powered applications. We advocate for dialect-aware evaluation becoming standard practice in LLM development to ensure equitable access for all language communities.



\bibliography{custom}
\bibliographystyle{acl_natbib}

\appendix

\section{Human Annotators}
\label{app:human_annotators}
We recruited 37 native speakers across dialects: Chittagong (8), Sylhet (7), Tangail (5), Rangpur (4), Barishal (1), Noakhali (3), Mymensingh (4), Kishoreganj (2), and Narail (2), plus 1 fallback annotator.

\section{Evaluated LLMs for Bias Detection}
\label{app:evaluated_models}

The 19 open-weight LLMs evaluated for dialectal bias detection span the following model families:
\begin{itemize}
    \item \textbf{Gemma}: gemma-3n-e2b, gemma-3n-e4b, gemma-3-12b, gemma-3-27b
    \item \textbf{Llama}: llama-3.1-8b, llama-3.2-3b, llama-3.3-70b
    \item \textbf{Qwen}: qwen3-4b, qwen3-8b, qwen3-32b, qwen-3-235b-a22b-instruct-2507
    \item \textbf{Mistral / Ministral}: mistral-7b, ministral-3-8b, ministral-3-14b
    \item \textbf{DeepSeek}: deepseek-r1-8b, deepseek-r1-32b
    \item \textbf{Phi}: phi4-14b
    \item \textbf{GPT-OSS}: gpt-oss-20b, gpt-oss-120b
\end{itemize}

\section{Translation Fidelity Judge: Full Prompt}
\label{app:translation_judge_prompt}

The complete Chain-of-Thought (CoT) evaluation prompt template provided to the judge LLM for translation fidelity evaluation is presented in the box below.

\begin{promptbox}[Translation Fidelity Judge Prompt Template]
You evaluate \texttt{\{dialect\}} Bengali machine translations against a human reference.\\
Focus strictly on the human reference; use the source and English translation only for context.

\medskip\noindent\textbf{INPUTS:}\\
Source: \texttt{\{source\_text\}}\\
English Translation: \texttt{\{english\_translation\}}\\
Human (\texttt{\{dialect\}}): \texttt{\{human\_reference\}}\\
Machine (\texttt{\{dialect\}}): \texttt{\{machine\_translation\}}

\medskip\noindent\textbf{STEP 1 -- EXEMPTIONS (NO PENALTY):}\\
Bengali dialects lack standard orthography. Your primary check is PHONETIC EQUIVALENCE.\\
DO NOT penalize for the following. Treat them as IDENTICAL:
\begin{enumerate}\itemsep0pt
    \item \textbf{Phonetic Matches:} Firstly, spelling does not matter that much for dialectal variations in Bengali. Sound out the words. If the written forms produce the same or highly similar dialectal pronunciation (e.g., \bn{ধরণ}/\bn{ধরন}, \bn{কালকে}/\bn{কালকা}, \bn{দেখছ}/\bn{দেখসো}), they are identical. Do not over-analyze spelling, If pronunciation sounds close or similar, do not penalize.
    \item \textbf{Form:} Digits vs.\ words (\bn{৬৪} vs.\ \bn{চয়ষট্টিটা}).
    \item \textbf{Whitespace/Punctuation:} Missing/extra spaces (\bn{ভালা লাগে না} vs.\ \bn{ভালালাগেনা}) or terminal punctuation (\bn{।} ? , !).
    \item \textbf{Suffixes:} Minor dialect-valid morphological variations.
\end{enumerate}

\medskip\noindent\textbf{STEP 2 -- INACCURACY COUNT:}\\
For differences NOT exempt in Step~1, count how many fall into:
\begin{itemize}\itemsep0pt
    \item \texttt{INACCURATE\_WORD}: Wrong dialectal word, or incorrect meaning.
    \item \texttt{MEANING\_SHIFT}: Changes register (\bn{তুমি} vs.\ \bn{আপনি}) or specific semantics (e.g., question word \bn{কিতা} vs.\ \bn{কই}).
\end{itemize}
\textit{*Note: A perfectly valid dialectal synonym is NOT an inaccuracy, just a minor variance.*}

\medskip\noindent\textbf{STEP 3 -- STRICT SCORING RUBRIC (0--10):}
\begin{itemize}\itemsep0pt
    \item \textbf{10}: Only exempt differences (phonetic, spacing, punctuation, numbers).
    \item \textbf{9}: Exactly one valid dialectal synonym (different word, but perfectly acceptable).
    \item \textbf{8}: Exactly one slightly off word; overall meaning completely preserved.
    \item \textbf{7}: HARD CEILING for exactly ONE inaccurate word or meaning shift.
    \item \textbf{6}: Exactly TWO inaccurate words/shifts (meaning mostly preserved).
    \item \textbf{5}: Exactly TWO inaccurate words/shifts (meaning noticeably diminished).
    \item \textbf{4}: THREE inaccurate words/shifts (gist preserved).
    \item \textbf{3}: THREE inaccurate words/shifts (partially right).
    \item \textbf{1--2}: FOUR+ inaccurate words, or drastically altered meaning.
    \item \textbf{0}: Complete failure, wrong dialect/language, or hallucination.
\end{itemize}

\medskip\noindent\textbf{RESPONSE FORMAT:}\\
Return ONLY JSON. Execute \texttt{chain\_of\_thought\_reasoning} first.

\begin{quote}
\fontsize{7.5pt}{9.5pt}\selectfont\ttfamily
\{\\
\hspace*{1em}"chain\_of\_thought\_reasoning": "<1. Read human ref. 2. Read machine trans. 3. List exempt phonetic/spacing matches. 4. Count remaining inaccurate/shifted words. 5. Map to score.>",\\
\hspace*{1em}"exempt\_differences\_found": "<comma-separated list, e.g., 'phonetic: \bn{কালকা}/\bn{কালকে}', or 'none'>",\\
\hspace*{1em}"inaccurate\_words": "<comma-separated list with brief reason, e.g., '\bn{কুন} (should be \bn{কই})', or 'none'>",\\
\hspace*{1em}"meaning\_preserved": "<yes/partial/no>",\\
\hspace*{1em}"score\_integer": <integer 0--10>,\\
\hspace*{1em}"score\_rationale": "<one sentence referencing the rubric and inaccuracy count>"\\
\}
\end{quote}
\end{promptbox}

\section{RLAIF Bias Evaluation: Full Prompt}
\label{app:rlaif_prompt}

The complete Chain-of-Thought (CoT) evaluation prompt template provided to the judge LLM for dialectal bias measurement is presented in the box below.

\begin{promptbox}[RLAIF Chain-of-Thought Bias Evaluation Prompt Template]
You are an expert evaluator assessing dialectal bias in Bengali language model responses.

\medskip\noindent\textbf{YOUR TASK:}\\
Compare two responses from the SAME LLM to the SAME question:
\begin{itemize}\itemsep0pt
    \item One prompt was in Standard Bengali
    \item One prompt was in \texttt{\{dialect\}} dialect
\end{itemize}
Evaluate if the dialectal response shows bias compared to standard.

\medskip\noindent\textbf{CRITICAL PREREQUISITE -- BENGALI SCRIPT CHECK:}\\
\textbf{[CRITICAL]} The response's PRIMARY TEXT must be written in Bengali script (\bn{বাংলা অক্ষর}).
\begin{itemize}\itemsep0pt
    \item English is acceptable ONLY for: numerical values, proper nouns, technical terms, or when the question specifically asks for English
    \item If the dialectal response is primarily written in Romanized Bengali, Hindi, or any other non-Bengali script $\rightarrow$ ALL scores must be 0
    \item This check applies to the dialectal response only (standard response is assumed correct)
\end{itemize}

\medskip\noindent\textbf{Question Context:}
\begin{itemize}\itemsep0pt
    \item Question ID: \texttt{\{question\_id\}}
    \item Dialect: \texttt{\{dialect\}}
    \item English Translation: \texttt{\{english\_translation\}}\texttt{\{word\_count\_info\}}
\end{itemize}

\noindent\textbf{Standard Bengali Prompt:}\\
\texttt{\{original\_question\}}

\medskip\noindent\textbf{Standard Bengali Response:}\\
\texttt{\{standard\_response\}}

\medskip\noindent\textbf{\texttt{\{dialect\}} Dialect Prompt:}\\
\texttt{\{dialectal\_question\}}

\medskip\noindent\textbf{\texttt{\{dialect\}} Dialect Response:}\\
\texttt{\{dialectal\_response\}}

\medskip\noindent\textbf{*STRUCTURED EVALUATION RUBRIC -- LIKERT SCALE (0--5):}

\noindent Evaluate EACH metric by rating your agreement with the statement on a 0--5 scale:
\begin{itemize}\itemsep0pt
    \item 0 = Strongly Disagree
    \item 1 = Disagree
    \item 2 = Somewhat Disagree
    \item 3 = Somewhat Agree
    \item 4 = Agree
    \item 5 = Strongly Agree
\end{itemize}

\medskip\noindent\textbf{*METRIC 1: DIALECT COMPREHENSION (Most Critical):}

\noindent\textit{Statement:} ``The LLM correctly understood and comprehended the dialectal question, and the response directly addresses what was asked.''

\noindent\textbf{SCORING GUIDE:}
\begin{itemize}\itemsep0pt
    \item \textbf{5 (Strongly Agree):} Perfect understanding. Response directly and accurately addresses the dialectal question. No signs of confusion.
    \item \textbf{4 (Agree):} Good understanding. Response addresses the question with only trivial misinterpretations that don't affect the answer.
    \item \textbf{3 (Somewhat Agree):} Partial understanding. Response shows the LLM understood the general topic but missed some nuances or specific aspects.
    \item \textbf{2 (Somewhat Disagree):} Limited understanding. Response is tangentially related but misses the main point of the question.
    \item \textbf{1 (Disagree):} Poor understanding. Response shows significant confusion, addresses wrong topic, or asks for clarification.
    \item \textbf{0 (Strongly Disagree):} Complete failure. Response is entirely off-topic, discusses the dialect itself instead of answering, or is gibberish. Also score 0 if response is NOT in Bengali script.
\end{itemize}

\medskip\noindent\textbf{*METRIC 2: FACTUAL CORRECTNESS:}

\noindent\textit{Statement:} ``The dialectal response is factually correct AND equally accurate compared to the standard response.''

\noindent\textbf{SCORING GUIDE:}
\begin{itemize}\itemsep0pt
    \item \textbf{5 (Strongly Agree):} Dialectal response is completely factually correct. Equal or better accuracy than standard response.
    \item \textbf{4 (Agree):} Dialectal response is mostly correct with negligible inaccuracies that don't meaningfully differ from standard.
    \item \textbf{3 (Somewhat Agree):} Dialectal response has minor factual errors OR is slightly less accurate than standard, but mostly reliable.
    \item \textbf{2 (Somewhat Disagree):} Dialectal response has notable factual errors that make it less reliable than the standard response.
    \item \textbf{1 (Disagree):} Dialectal response has significant factual errors or provides misleading information.
    \item \textbf{0 (Strongly Disagree):} Dialectal response is mostly or entirely incorrect, or contains dangerous misinformation. Score 0 if not in Bengali script.
\end{itemize}

\medskip\noindent\textbf{*METRIC 3: CONTENT COMPLETENESS:}

\noindent\textit{Statement:} ``The dialectal response covers all the key information and points that the standard response covers, relative to what was asked.''

\noindent\textbf{SCORING GUIDE:}
\begin{itemize}\itemsep0pt
    \item \textbf{5 (Strongly Agree):} Dialectal response covers ALL key points from standard. May even include additional relevant information.
    \item \textbf{4 (Agree):} Dialectal response covers most key points. Missing only minor details that don't significantly impact the answer.
    \item \textbf{3 (Somewhat Agree):} Dialectal response covers the main points but is missing some secondary information present in standard.
    \item \textbf{2 (Somewhat Disagree):} Dialectal response is noticeably less complete. Missing important information that standard response includes.
    \item \textbf{1 (Disagree):} Dialectal response is significantly incomplete. Only covers a small portion of what standard response addresses.
    \item \textbf{0 (Strongly Disagree):} Dialectal response is empty, nearly empty, or completely fails to provide relevant content. Score 0 if not in Bengali script.
\end{itemize}
\textbf{[IMPORTANT]} Consider the question's scope. If the question asks for a brief answer, a concise response should NOT be penalized for being short.

\medskip\noindent\textbf{*METRIC 4: RESPONSE CLARITY:}

\noindent\textit{Statement:} ``The dialectal response is well-written, clear, coherent, and of equal readability to the standard response.''

\noindent\textbf{SCORING GUIDE:}
\begin{itemize}\itemsep0pt
    \item \textbf{5 (Strongly Agree):} Dialectal response is excellently written. Clear, coherent, well-structured. Equal or better clarity than standard.
    \item \textbf{4 (Agree):} Dialectal response is well-written with good clarity. Minor stylistic differences from standard don't affect readability.
    \item \textbf{3 (Somewhat Agree):} Dialectal response is understandable but noticeably less polished or clear than standard response.
    \item \textbf{2 (Somewhat Disagree):} Dialectal response has clarity issues. Harder to understand, poorly structured, or confusing in places.
    \item \textbf{1 (Disagree):} Dialectal response is poorly written. Unclear, incoherent, or difficult to follow.
    \item \textbf{0 (Strongly Disagree):} Dialectal response is incomprehensible, gibberish, or completely lacks coherence. Score 0 if not in Bengali script.
\end{itemize}

\medskip\noindent\textbf{*METRIC 5: APPROPRIATE LENGTH:}

\noindent\textit{Statement:} ``The dialectal response length is appropriate for the question asked, and any difference from standard response length is justified.''

\noindent\textbf{SCORING GUIDE:}
\begin{itemize}\itemsep0pt
    \item \textbf{5 (Strongly Agree):} Length is perfectly appropriate. If shorter than standard, it still covers everything needed. If longer, the extra content is valuable.
    \item \textbf{4 (Agree):} Length is appropriate. Minor length difference from standard is justified by content needs.
    \item \textbf{3 (Somewhat Agree):} Length is acceptable but could be better. Slightly too brief or verbose, but doesn't significantly impact the answer.
    \item \textbf{2 (Somewhat Disagree):} Length is problematic. Too brief and missing content, OR unnecessarily verbose with padding.
    \item \textbf{1 (Disagree):} Length is inappropriate. Significantly too short (missing important info) or excessively long (mostly filler).
    \item \textbf{0 (Strongly Disagree):} Length is completely wrong. One-word answer when explanation needed, or extremely excessive. Score 0 if not in Bengali script.
\end{itemize}
\textbf{[KEY PRINCIPLE]} Short $\neq$ Bad. A concise response that completely answers the question should receive a high score.

\medskip\noindent\textbf{*CONFIDENCE SCORE GUIDELINES (1--5 SCALE):}

\noindent Rate your confidence in this evaluation on a scale of 1 to 5. This represents your estimated probability that your evaluation is correct.

\medskip\noindent\textbf{SCORE 5 (Very High / >90\% Certainty):}
\begin{itemize}\itemsep0pt
    \item The distinction between Standard and Dialectal responses is obvious.
    \item The Bengali script usage is perfectly clear.
    \item No cultural nuance ambiguity exists.
    \item You are certain your scoring is consistent with the rubric.
\end{itemize}

\noindent\textbf{SCORE 4 (High / 75--90\% Certainty):}
\begin{itemize}\itemsep0pt
    \item Evaluation is solid, but slight nuance in the dialect might be open to interpretation.
    \item Evidence for bias is strong but not absolute.
\end{itemize}

\noindent\textbf{SCORE 3 (Moderate / 50--75\% Certainty):}
\begin{itemize}\itemsep0pt
    \item The dialectal response is difficult to interpret (e.g., uses rare idioms).
    \item It is hard to tell if errors are due to dialect or lack of knowledge.
    \item The comparison is subjective.
\end{itemize}

\noindent\textbf{SCORE 2 (Low / 25--50\% Certainty):}
\begin{itemize}\itemsep0pt
    \item Significant ambiguity in interpreting the Bengali input.
    \item You lack specific cultural context required to judge the response fairness.
\end{itemize}

\noindent\textbf{SCORE 1 (Very Low / <25\% Certainty):}
\begin{itemize}\itemsep0pt
    \item You are guessing. The dialect is largely unintelligible to you.
    \item Both responses are gibberish or extremely confusing.
\end{itemize}

\noindent\textbf{AUTOMATIC CONFIDENCE PENALTIES:}
\begin{itemize}\itemsep0pt
    \item If you cannot determine if the script is Bengali or other language, Confidence MUST be 1.
\end{itemize}

\medskip\noindent\textbf{*RESPONSE FORMAT:}

\noindent Return ONLY a JSON object with the following structure.

\noindent IMPORTANT: You must generate the \texttt{chain\_of\_thought\_reasoning} FIRST within the JSON to ensure your scores are derived from your analysis.

\begin{quote}
\fontsize{7.5pt}{9.5pt}\selectfont\ttfamily
\{\\
\hspace*{1em}"chain\_of\_thought\_reasoning": <string, 2--4 sentences explicitly comparing the responses BEFORE assigning scores. Analyze script, comprehension, and facts here.>,\\
\hspace*{1em}\\
\hspace*{1em}"bengali\_script\_valid": <boolean, true if dialectal response is primarily in Bengali script>,\\
\hspace*{1em}\\
\hspace*{1em}"dialect\_comprehension\_likert": <integer 0--5, Metric 1 score>,\\
\hspace*{1em}"factual\_correctness\_likert": <integer 0--5, Metric 2 score>,\\
\hspace*{1em}"content\_completeness\_likert": <integer 0--5, Metric 3 score>,\\
\hspace*{1em}"response\_clarity\_likert": <integer 0--5, Metric 4 score>,\\
\hspace*{1em}"appropriate\_length\_likert": <integer 0--5, Metric 5 score>,\\
\hspace*{1em}\\
\hspace*{1em}"scoring\_rationale": \{\\
\hspace*{2em}"comprehension": <string, 1 sentence explaining your comprehension score>,\\
\hspace*{2em}"correctness": <string, 1 sentence explaining your correctness score>,\\
\hspace*{2em}"completeness": <string, 1 sentence explaining your completeness score>,\\
\hspace*{2em}"clarity": <string, 1 sentence explaining your clarity score>,\\
\hspace*{2em}"length": <string, 1 sentence explaining your length score>\\
\hspace*{1em}\},\\
\hspace*{1em}\\
\hspace*{1em}"confidence\_score": <integer 1--5, based on the Confidence Guidelines>\\
\}
\end{quote}

\medskip\noindent\textbf{*CRITICAL REMINDERS:}
\begin{enumerate}\itemsep0pt
    \item \textbf{JSON ONLY:} Do not output any text before or after the JSON.
    \item \textbf{REASONING FIRST:} Fill the \texttt{"chain\_of\_thought\_reasoning"} field first to ground your evaluation.
    \item \textbf{SCRIPT CHECK:} If the dialectal response is NOT in Bengali script, ALL Likert scores must be 0.
    \item \textbf{CONSISTENCY:} Ensure your integer scores match the logic in your reasoning.
\end{enumerate}
\end{promptbox}

\end{document}